\documentclass[letterpaper, 10 pt, conference]{ieeeconf}  % 

\IEEEoverridecommandlockouts                              % This command is only needed if 

\usepackage{graphics} % for pdf, bitmapped graphics files
\usepackage{epsfig} % for postscript graphics files
\usepackage{times} % assumes new font selection scheme installed
\usepackage{amsmath} % assumes amsmath package installed
\usepackage{amssymb}  % assumes amsmath package installed
\usepackage{bm}
\usepackage{color}
\usepackage{cite}
\usepackage[linesnumbered,ruled]{algorithm2e}
\usepackage{ulem} %to strike the words
\usepackage{hyperref}
\usepackage{float}
\usepackage{booktabs}
\usepackage{multirow}
\usepackage{mathrsfs}
\usepackage[utf8]{inputenc}
\usepackage[caption=false,font=footnotesize]{subfig}
\usepackage{balance}

\usepackage{graphicx}
\usepackage{pifont}
\usepackage{soul}

\newcounter{RNum}
\renewcommand{\theRNum}{\arabic{RNum}}
\newcommand{\Remark}{\noindent\textit{\textbf{Remark}~\refstepcounter{RNum}\textbf{\theRNum}:}}
\newcommand{\NoOne}[1]{\textcolor{red}{#1}}
\newcommand{\NoTwo}[1]{\textcolor{green}{#1}}
\newcommand{\NoThree}[1]{\textcolor{blue}{#1}}
\soulregister\NoOne7
\soulregister\NoTwo7
\soulregister\NoThree7
\soulregister\Remark7

\title{\LARGE \bf
	Siamese Object Tracking for Vision-Based UAM Approaching with Pairwise Scale-Channel Attention
	
}

\author{Guangze Zheng$^1$, Changhong Fu$^{1,*}$, Junjie Ye$^1$, Bowen Li$^1$, Geng Lu$^2$, and Jia Pan$^3$
	\thanks{*Corresponding Author}% <-this % stops a space
	\thanks{$^1$G. Zheng, C. Fu, J. Ye, and B. Li are with the School of Mechanical Engineering, Tongji University, Shanghai 201804, China.
		{\tt\small changhongfu@tongji.edu.cn}}
	\thanks{$^2$G. Lu is with the Department of Automation, Tsinghua University, Beijing 100084, China.}
	\thanks{$^3$J. Pan is with the Department of Computer Science, the University of Hong Kong, Hong Kong, China}
}

\begin{document}
	\maketitle
	\thispagestyle{empty}
	\pagestyle{empty}
	
	%%%%%%%%%%%%%%%%%%%%%%%%%%%%%%%%%%%%%%%%%%%%%%%%%%%%%%%%%%%%%%%%%%%%%%%%%%%%%%%%%
	%%%%%%%%%%%%%%%%%%%%%%%%%%%%%%%%%%     Abstract     %%%%%%%%%%%%%%%%%%%%%%%%%%%%%
	%%%%%%%%%%%%%%%%%%%%%%%%%%%%%%%%%%%%%%%%%%%%%%%%%%%%%%%%%%%%%%%%%%%%%%%%%%%%%%%%%
	
	\begin{abstract}
		%Visual approaching the object is crucial to the subsequent manipulating of the unmanned aerial manipulator (UAM).
		%Although the manipulating has been widely studied, the vision-based UAM approaching generally lacks efficient design. 
		Although the manipulating of the unmanned aerial manipulator (UAM) has been widely studied, vision-based UAM approaching, which is crucial to the subsequent manipulating, generally lacks effective design.
		The key to the visual UAM approaching lies in object tracking, while current UAM tracking typically relies on costly model-based methods. Besides, UAM approaching often confronts more severe object scale variation issues, which makes it inappropriate to directly employ state-of-the-art model-free Siamese-based methods from the object tracking field. 
		To address the above problems, this work proposes a novel Siamese network with pairwise scale-channel attention (SiamSA) for vision-based UAM approaching. Specifically, SiamSA consists of a pairwise scale-channel attention network (PSAN) and a scale-aware anchor proposal network (SA-APN). PSAN acquires valuable scale information for feature processing, while SA-APN mainly attaches scale awareness to anchor proposing.
		Moreover, a new tracking benchmark for UAM approaching, namely UAMT100, is recorded with 35K frames on a flying UAM platform for evaluation. Exhaustive experiments on the benchmarks and real-world tests validate the efficiency and practicality of SiamSA with a promising speed. Both the code and UAMT100 benchmark are now available at \url{https://github.com/vision4robotics/SiamSA}.

	\end{abstract}
	
	%%%%%%%%%%%%%%%%%%%%%%%%%%%%%%%%%%%%%%%%%%%%%%%%%%%%%%%%%%%%%%%%%%%%%%%%%%%%%%%%%
	%%%%%%%%%%%%%%%%%%%%%%%%%%%%%%%%%    Introduction   %%%%%%%%%%%%%%%%%%%%%%%%%%%%%
	%%%%%%%%%%%%%%%%%%%%%%%%%%%%%%%%%%%%%%%%%%%%%%%%%%%%%%%%%%%%%%%%%%%%%%%%%%%%%%%%%

	\section{Introduction}
	%For multiple practical applications~\cite{perch, wristband, 6943038,  UAMreview,7139760,5980314,aerialpick},
	For multiple practical scenarios~\cite{perch, wristband, 6943038, aerialpick},
	\textit{e.g.}, autonomous grasping~\cite{perch}, aerial pick-and-place~\cite{aerialpick}, and wristband placement~\cite{wristband}, the applications of unmanned aerial manipulator (UAM) mainly consist of two stages, \textit{i.e.}, visual approaching and precisely manipulating the object.
	The key to visual UAM approaching lies in the object tracking method, which provides the continuous visual perception of the object.
	However, current tracking methods for vision-based UAM approaching are generally model-based~\cite{wristband}\cite{aerialpick}\cite{delivery}. 
	These methods commonly rely on predefined class labels and require specific large-scale training datasets, which are inadequate to track arbitrary objects and increase heavy workload for source-limited UAM platforms.
	Besides, collecting specific datasets for training model-based methods also causes huge development costs. 
	An intuitive solution is to introduce the model-free object tracking methods for UAM approaching. 
	In the visual object tracking field, Siamese network-based methods~\cite{SiamRPN++, SiamFC++, SiamAPN,UDAT,tctrack} have surpassed correlation filter-based methods~\cite{TIE,MSCF, Autotrack, grsm, ARCF} and reached state-of-the-art (SOTA) performance. 
	The model-free Siamese trackers can perform on any online-assigned object to meet the requirements of practical application with an affordable camera. 
	Therefore, this work introduces Siamese network as the object tracking method for vision-based UAM approaching.
	
	\begin{figure}[t]
		\setlength{\abovecaptionskip}{-0.2cm}
		\centering
		\includegraphics[width=0.42\textwidth]{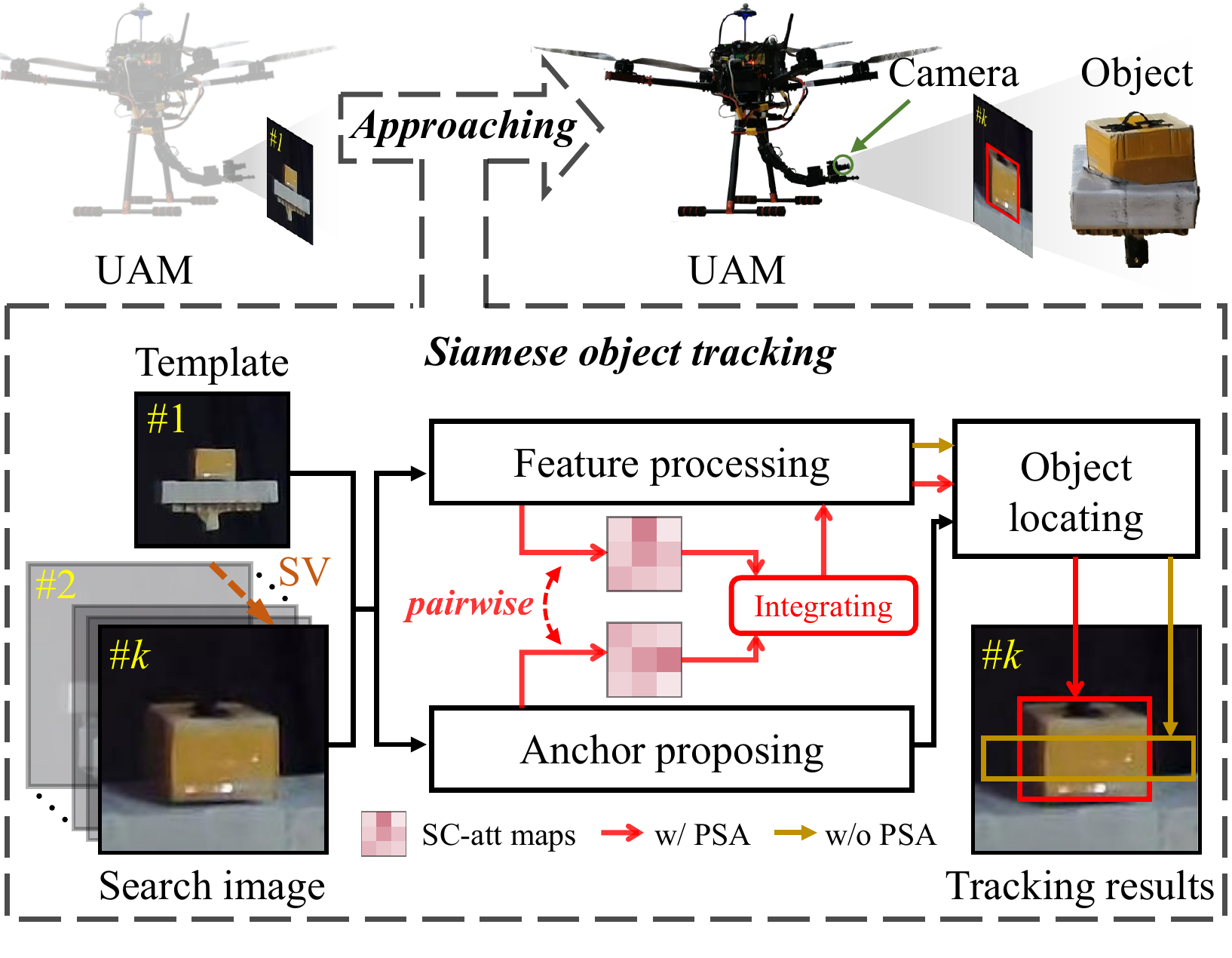}
		\caption
		{A UAM approaching process and qualitative comparison. The results demonstrate the effects of the proposed pairwise scale-channel attention (PSA). 
			The result in \textcolor[rgb]{ 1,  0, 0}{red} box shows that tracking with PSA is more robust to deal with severe scale variation (SV) in UAM approaching than without PSA (\textcolor[rgb]{ 0.749,  0.565, 0}{brown} box).
		}
		%\vspace{-0cm}
		\label{fig:fig1}
	\end{figure}

	\begin{figure*}[t]	
		\centering
		\includegraphics[width=0.85\linewidth]{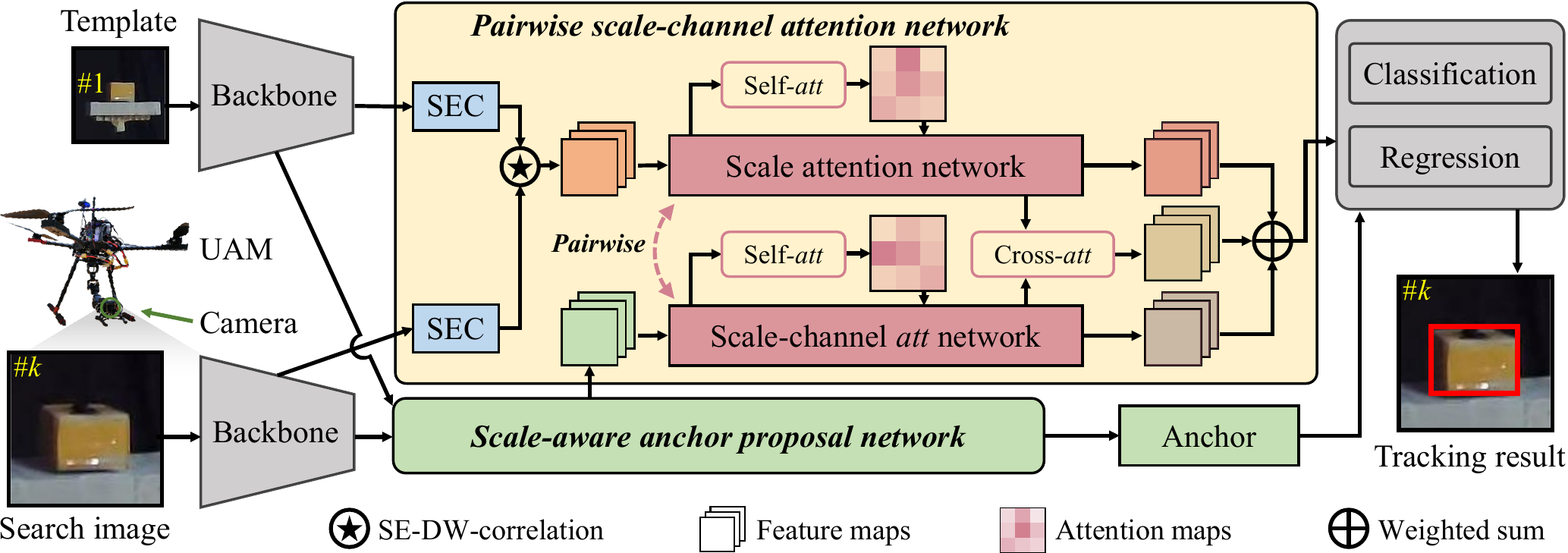}
		\setlength{\abovecaptionskip}{-0.1cm}
		\caption
		{An overview of the proposed Siamese tracking with pairwise scale-channel attention (SiamSA) for UAM approaching. SEC refers to scale-equivariant convolution, and \textit{att} denotes attention.
		}  
		\label{fig:main}
		\vspace{-0.5cm}
	\end{figure*}

	Another issue of vision-based UAM approaching is the extreme object scale variation (SV), which is rarely taken into account. From the perspective of the UAM onboard camera, the scale of objects will get larger as the UAM approaches it.
	Because the distance between the UAM and the object is usually limited in practical applications, even modest changes in relative distance can result in significant scale variation, posing a major barrier for general object tracking methods. 
	To address the SV problem, this work uses scale-equivariant convolution as the first step.
	In addition, attention-based strategies are elaborately designed to further solve SV issues. Because of the success in feature refinement~\cite{SiamAttn}\cite{RASNet}, these strategies have sparked considerable concern. In general, attention-based methods infer attention maps among spatial and channel dimensions, while this work attempts to uncover the scale attention for UAM approaching tasks. 
	Specifically, a novel pairwise scale-channel attention (PSA) is proposed to extract important scale information across channels. As shown in Fig.~\ref{fig:fig1}, in both feature processing and anchor proposing, scale-channel attention (SC-att) maps are inferred in a pairwise structure and integrated to locate the object with stronger scale awareness. 
	
	Two branches are built based on the proposed PSA: a pairwise scale-channel attention network (PSAN) and a scale-aware anchor proposal network (SA-APN). PSAN seeks to express more resilient features against extreme object scale variation, whereas SA-APN links scale information to anchors in the anchor proposal network. Self-attention and cross-attention tactics are used.
	
	A fair evaluation of object tracking methods is required for vision-based UAM approaching. Despite vision-based UAM approaching being critical for practical applications, there are no publicly available benchmarks to assess UAM tracking methods for approaching yet.
	Therefore, this work records UAMT100 on a flying UAM platform with 100 image sequences.
	The UAMT100 benchmark covers common object tracking challenges and also introduces the unique challenge of UAM approaching, \textit{i.e.}, UAM attack.
	The videos are taken in an indoor environment with a motion-capture system. 
	SiamSA is also evaluated on the challenging authoritative UAV tracking benchmark to verify the generality of the proposed pairwise scale awareness. Finally, the proposed SiamSA has been deployed on a UAM platform and has proven to be highly efficient.
	The main contributions of this work can be summarized as follows:
	\begin{itemize}
		\item A Siamese object tracking method with novel pairwise scale-channel attention (SiamSA) is proposed for vision-based UAM approaching.
		\item A new pairwise scale-channel attention network and a scale-aware anchor proposal network are proposed to solve the scale variation issues in UAM approaching.
		\item A novel UAMT100 benchmark with precise annotations is built for evaluating vision-based UAM tracking methods for approaching. 
		\item Exhaustive experiments on the new UAMT100 benchmark, the authoritative aerial tracking benchmark, and real-world tests have verified the practicality and effectiveness of SiamSA for efficient UAM approaching.
	\end{itemize}

	%%%%%%%%%%%%%%%%%%%%%%%%%%%%%%%%%%%%%%%%%%%%%%%%%%%%%%%%%%%%%%%%%%%%%%%%%%%%%%%%%
	%%%%%%%%%%%%%%%%%%%%%%%%%%%%%%%%%    Related Work   %%%%%%%%%%%%%%%%%%%%%%%%%%%%%
	%%%%%%%%%%%%%%%%%%%%%%%%%%%%%%%%%%%%%%%%%%%%%%%%%%%%%%%%%%%%%%%%%%%%%%%%%%%%%%%%%
	%	%\vspace{-4pt}
	\section{Related Works}
	
	%	\vspace{-4pt}
	\subsection{Object tracking for UAM approaching}
	%	\vspace{-2pt}
	In numerous practical vision-based UAM approaching scenarios, continuous perception of the object location has been critical, resulting in urgent demand for tracking methods' efficiency and practicability.
	%The requirement of UAM tracking originates from some practical applications.
	G. Garimella~\textit{et al.}~\cite{aerialpick}
	%	 designed an onboard system and 
	use a detection method for UAM tracking when dealing with aerial pick-and-place. However, the detection method relies on LED markers, which has constrained the automation of UAM.
	During the research on picking and delivery of magnetic objects, A. Gawel~\textit{et al.}~\cite{delivery} perform object detection instead of tracking methods to track the object's center of gravity. They also find the deficiency of detection methods for UAM approaching and prepare to implement object tracking. Similarly, J. M. G\'omez-de Gabriel~\textit{et al.}~\cite{wristband}
	%the authors in~\cite{wristband} 
	adopt an object detection method on UAM in the wristband placement task. Such kind of detection methods depends on predefined class labels, which cannot track various kinds of object and brings inflexibility. Besides, collecting specific datasets for training model-based methods also causes enormous extra development costs. 
	To better accomplish UAM tracking requirements and promote automation to a higher degree, this work introduces the model-free Siamese tracking methods for UAM approaching.
	%	\vspace{-4pt}
	\subsection{Siamese tracking for aerial systems}
	%	\vspace{-2pt}
	Model-free Siamese tracking has been prevalent in the object tracking field due to its SOTA efficiency and robustness. %Siamese-based tracking methods are generally divided into anchor-based and anchor-free methods.
	%As an anchor-based method, 
	SiamRPN~\cite{SiamRPN} introduces region proposal network (RPN) into Siamese tracking to increase precision and robustness with anchors. SiamRPN++~\cite{SiamRPN++} employs deeper features and further improves effectiveness.  %Anchor-free strategies, \textit{e.g.},
	SiamFC++~\cite{SiamFC++} abandons anchors by regressing offsets to reduce hyper-parameters and classification errors.
	Afterward, SiamAPN~\cite{SiamAPN} improves Siamese network for UAV tracking,
	%Combining the advantages of both strategies, SiamAPN  
	which streamlines anchor generation strategy by regressing offsets, thereby reducing hyper-parameters for UAV tracking with high efficiency. However, the scale variation issue in UAM approaching has posed a formidable challenge for directly employing Siamese tracking and requires an urgent solution.
	On the other hand, attention-based approaches have been widely used to capture specialized information in object tracking. These methods can tell where to focus inside the region of interest while also improving the tracking object's representation. RASNet~\cite{RASNet} designs residual, channel, and general attention modules to learn corresponding information. SiamAttn~\cite{SiamAttn} introduces deformable self-attention and cross-attention to learn context information and contextual inter-dependencies. These methods mainly infer spatial or channel attention maps, while this study focuses on scale variation issues and proposes scale attention for UAM approaching.

	%%%%%%%%%%%%%%%%%%%%%%%%%%%%%%%%%%%%%%%%%%%%%%%%%%%%%%%%%%%%%%%%%%%%%%%%%%%%%%%%%
	%%%%%%%%%%%%%%%%%%%%%%%%%%%%%%%%%%    Methodology   %%%%%%%%%%%%%%%%%%%%%%%%%%%%%
	%%%%%%%%%%%%%%%%%%%%%%%%%%%%%%%%%%%%%%%%%%%%%%%%%%%%%%%%%%%%%%%%%%%%%%%%%%%%%%%%%

	\begin{figure}[t]	
		
		\includegraphics[width=0.44\textwidth]{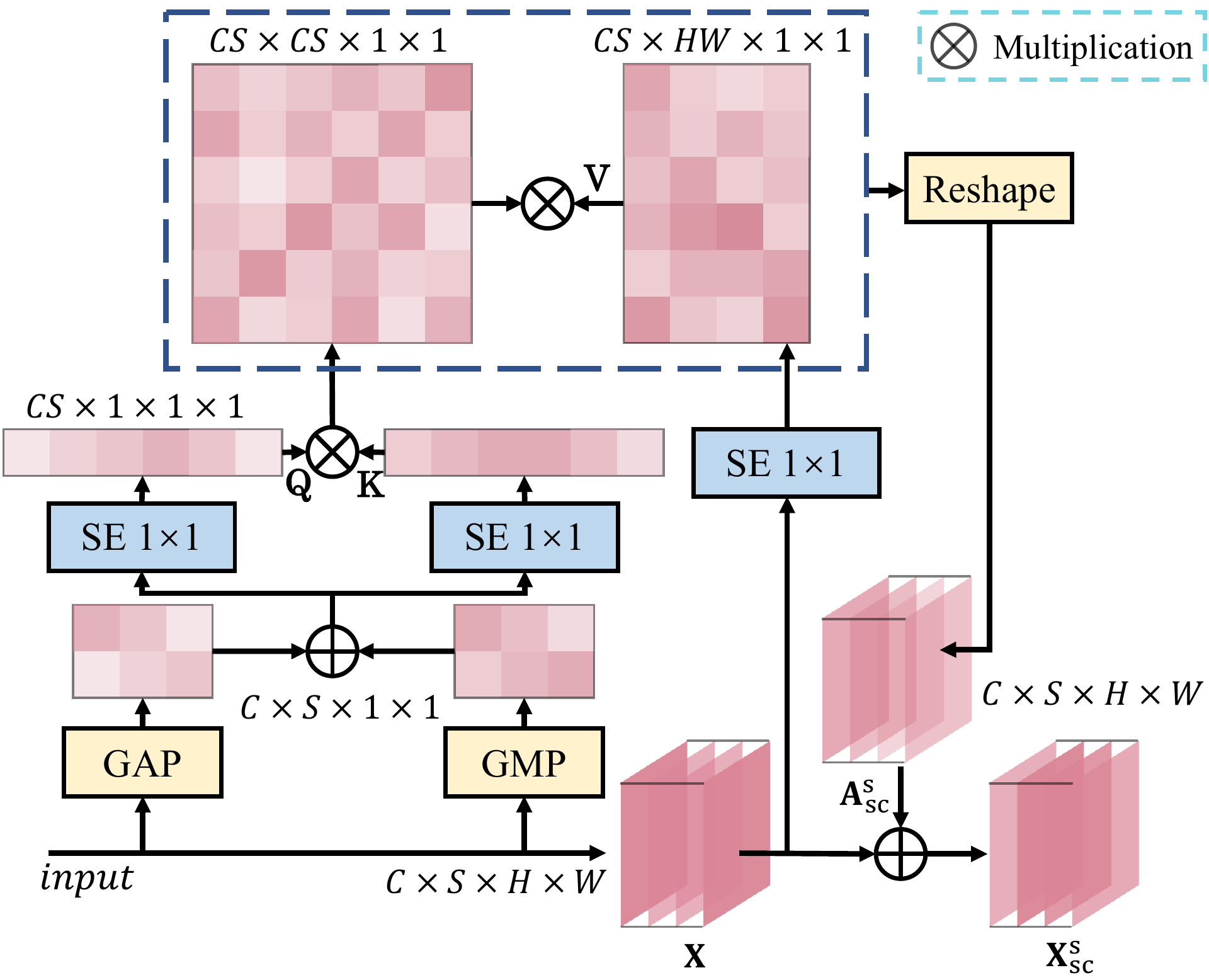}
		\centering
		\setlength{\abovecaptionskip}{0cm}
		%\vspace{-0.2cm}
		\caption
		{
			Main workflow of scale-channel self-attention strategy. The refined features are attached with scale information to better estimate object scale. 
		}
		\vspace{-0.1cm}
		\label{fig:att}
	\end{figure}
	
	\section{Proposed Method}
	
	The detailed introduction of SiamSA is mainly divided into three parts. First, scale-equivariant convolution is briefly reviewed. Afterward, the pairwise scale-channel attention network (PSAN) and the scale-aware anchor proposal network (SA-APN) are discussed in detail respectively. 
	Figure~\ref{fig:main} demonstrates an overview of the proposed SiamSA method %object tracking 
	for UAM approaching. 
	
	\subsection{Scale-equivariant convolution}
	Scale-equivariant (SE) convolution~\cite{scaletrack} in SiamSA enables UAM object tracking with preliminary object scale awareness for approaching. Common convolutional neural networks (CNNs) lack corresponding techniques for SV difficulties. However, SE convolution can tackle the problem with high computational efficiency, as described below:
	\begin{equation}
		%\footnotesize
		[f\star_{H}\kappa_\sigma](s,t)=\sum_{s'}[f(s',\cdot)\star\kappa_{s\cdot\sigma}(s^{-1}s',\cdot)](t)
		\quad,
	\end{equation}
	where $f\star_{H}(s,t)$ is a function of scale $s$ and translation $t$. $\kappa_\sigma(s,t)$ stands for a kernel. Besides, $\star$ represents common convolution operation, and $s'$ represents SV degree, where $s'>1$ means upscale while $s'<1$ denotes downscale. In this step, a preliminary perception of object scale is obtained by adding an additional scale dimension to image features.
	%In Fig.~\ref{fig:main}, as feature maps are extracted from backbone, scale-equivariant convolution generates an additional dimension to exploit scale information. 
	
	\noindent\textbf{\textit{Remark 1}}: Based on depthwise correlation in~\cite{SiamRPN++}, correlation is improved as SE-DW-correlation for UAM tracking.
	
	\subsection{Pairwise scale-channel attention network}
	PSAN aims to represent scale-aware feature maps during severe scale variation in UAM approaching. 
	%pairwise scale-channel attention 
	PSA utilizes correlation results and feature maps from SA-APN to excavate object scale information in PSAN, as shown in Fig.~\ref{fig:main}. %Both self-attention and cross-attention are performed to represent robust scale-aware features. 
	
	%	The PSAN mainly consists of two scale-channel attention networks that work in pairs,  
	%	Feature maps from the correlation and SA-APN feed into the pairwise scale-channel attention networks, respectively. 
	
	\begin{figure}[t]	
		\includegraphics[width=0.46\textwidth]{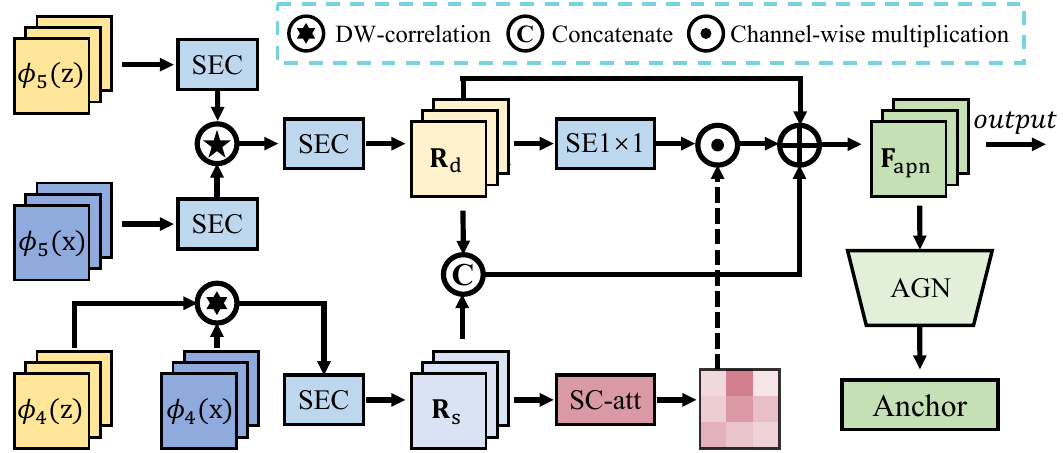}
		\centering
		\setlength{\abovecaptionskip}{-0.1cm}
		\caption
		{
			Main framework of scale-aware anchor proposal network (SA-APN). The output feature maps are sent to form pairwise scale-channel attention, and the generated anchors are equipped with scale awareness.
		}
		\vspace{-0.05cm}
		\label{fig:SA-APN}
	\end{figure}

	%\subsubsection{SC-att} 
	Given input scaled feature maps $\textbf{X}\in{\mathbb{R}^{C\times{S}\times{H}\times{W}}}$ from the SE-DW-correlation, where superscript $S$ represents the additional dimension related to scale,
	scale-channel self-attention is first employed. 
	B. Li~\textit{et al.}~\cite{SiamRPN++} demonstrate that a single channel in feature maps may focus on a specific class of objects, while the proposed SC self-attention takes advantage of this observation, and exploits scale information among different channels for stronger scale awareness. 
	Figure~\ref{fig:att} demonstrates the general pipeline of SC self-attention. 
	First, through the operation of global average pooling (GAP) and global max pooling (GMP) respectively, the input 4-dimensional features $\mathbf{X}\in{\mathbb{R}^{C\times{S}\times{H}\times{W}}}$ are turned into 2-dimension. Afterward, 2-dimensional features pass a fast 1×1 scale convolution layer and are flattened to 1-dimension to be query $\mathbf{Q}$ and key $\mathbf{K}$. $\mathbf{Q}$ is generated as:
	\begin{equation}
		\mathbf{Q}=\mathrm{Flat}([f\star_{H}\kappa'_\sigma](s, \mathrm{GAP}(\mathbf{X}); \mathrm{GMP}(\mathbf{X})))
		\quad,
	\end{equation}
	where $\mathbf{Q}\in{\mathbb{R}^{CS}}$, and $[f\star_{H}\kappa'_\sigma]$ denotes fast 1×1 scale convolution. Value $\mathbf{V}$ is acquired directly from the feature maps with fast 1×1 scale convolution, where value is reshaped to 2-dimension $\mathbf{V}\in{\mathbb{R}^{CS\times{HW}}}$. 
	Therefore, the scale-channel self-attention $\mathbf{A}^\mathrm{s}_\mathrm{sc}$ is represented as:
	\begin{equation}
		\mathbf{A}^\mathrm{s}_\mathrm{sc}(\mathbf{Q}, \mathbf{K}, \mathbf{V})=\mathrm{Softmax}(\mathbf{Q}\mathbf{K}^\top)\mathbf{V}
		\quad,
	\end{equation}
	where the superscript $\mathrm{s}$ denotes self-attention.
	
	The generated scale-channel self-attention refines the feature maps following the formula below:
	\begin{equation}\label{equ: feature}
		\mathbf{X}^\mathrm{s}_\mathrm{sc}=\mathbf{X} + \gamma^\mathrm{s}\mathbf{A}^\mathrm{s}_\mathrm{sc}
		\quad,	
	\end{equation}
	where $\mathbf{X}^\mathrm{s}_\mathrm{sc}\in{\mathbb{R}^{C\times{S}\times{H}\times{W}}}$ represents feature maps with scale-channel self-attention and $\gamma$ denotes a weight.
	
	\noindent\textit{\textbf{Remark 2}}: The same operations are performed on feature maps from both correlation results and SA-APN in a pairwise structure, exploiting scale information from both aspects.
	
	%During UAM approaching, the object scale changes significantly from the original appearance. Consequently,
	For further learning significant scale clues, the internal relationship between feature maps from SE-DW-correlation and SA-APN is worth learning by the cross-attention strategy. 
	%to obtain robust scale awareness.
	As scale-channel self-attention are generated, query $\mathbf{Q}_c\in{\mathbb{R}^{CS}}$ , key $\mathbf{K}_c\in{\mathbb{R}^{CS}}$ from the correlation and value $\mathbf{V}_a\in{\mathbb{R}^{CS\times{hw}}}$ from SA-APN are ready for cross-attention:
	\begin{equation}
		\mathbf{A}^\mathrm{c}_\mathrm{sc}(\mathbf{Q}_c, \mathbf{K}_c, \mathbf{V}_a)=\mathrm{Softmax}(\mathbf{Q}_c\mathbf{K}_c^\top)\mathbf{V}_a
		\quad,
	\end{equation}
	where superscript $\mathrm{c}$ means cross-attention. Similar to Eq.~\ref{equ: feature}, the scale-channel cross-attention $\mathbf{A}^\mathrm{c}_\mathrm{sc}\in{\mathbb{R}^{CS\times{hw}}}$ assists in the following generation of the refined feature maps $\mathbf{X}^\mathrm{c}_\mathrm{sc}$:
	\begin{equation}
		\mathbf{X}^\mathrm{c}_\mathrm{sc}=\mathbf{X} + \gamma^\mathrm{c}\mathbf{A}^\mathrm{c}_\mathrm{sc}
		\quad,	
	\end{equation}
	where $\mathbf{X}^\mathrm{c}_\mathrm{sc}\in{\mathbb{R}^{C\times{S}\times{H}\times{W}}}$ refers to feature maps from the SA-APN with scale-channel cross-attention related to PSAN.
	
	\noindent\textit{\textbf{Remark 3}}: Cross-attention integrates the useful scale information%from PSA%pairwise scale-channel attention
	, achieving better estimation of the object scale for UAM approaching.
	
	\begin{figure}[t]	
		%	%\vspace{-12pt}
		\includegraphics[width=0.47\textwidth]{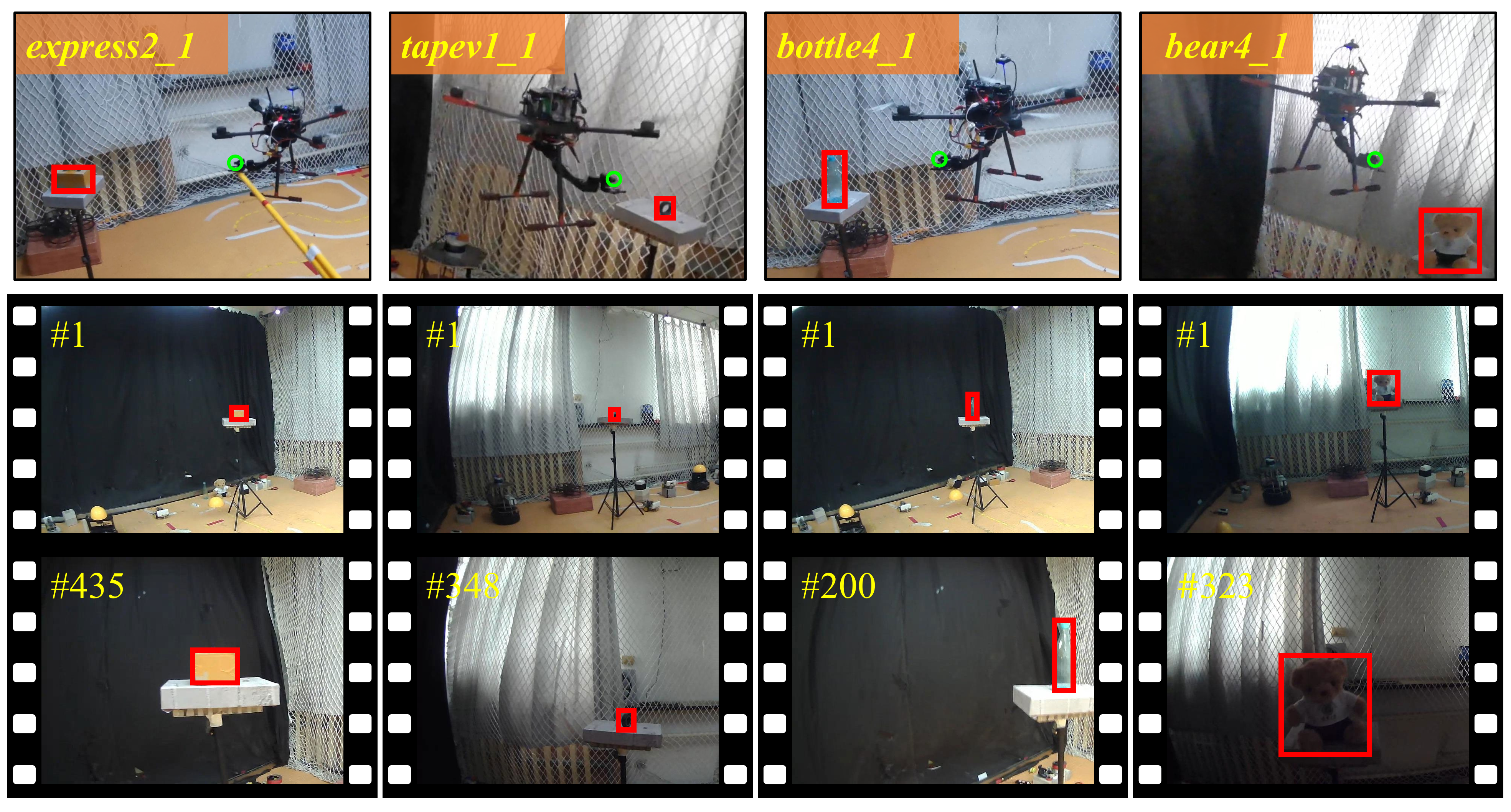}
		\centering
		\setlength{\abovecaptionskip}{-0cm}
		\caption
		{
			Examples of scenes from UAMT100. The first row is from a fixed camera % as the third perspective, 
			while the others are from the UAM onboard camera.
			%as the first perspective. 
			The scale of the objects suffers drastic variation during UAM approaching. The \textcolor[rgb]{ 1,  0, 0}{red} box and \textcolor[rgb]{ 0,  1, 0}{green} circle denote the object and the onboard camera, respectively.
		}
		%		\setlength{\abovecaptionskip}{-10pt}
		%		%\vspace{-0.4cm}
		\label{fig:dataset}
		\vspace{-0.1cm}
	\end{figure}
	
	%\vspace{-4pt}
	\subsection{Scale-aware anchor proposal network}
	%\vspace{-2pt}
	To attach meaningful scale information to anchor proposing, the scale-channel cross-attention strategy is also employed to form a scale-aware anchor proposal network (SA-APN). As shown in Fig.~\ref{fig:SA-APN}, %$\phi_5(\mathrm{z})$ represents the template features extracted from the Conv5 block of backbone, 
	$\phi_\mathrm{i}$ denotes the backbone layer corresponding to the subscript $\mathrm{i}\in\{\mathrm{4}, \mathrm{5}\}$, $\mathrm{x}$ refers to the search region, $\mathrm{z}$ points to the template, and AGN denotes the anchor generation network, which consists of two convolution layers.  Since the feature maps from shallower layers generally represent meaningful spatial information while the ones from deeper layers contain richer semantic clues, the scale-channel cross-attention is adopted to aggregate the information for stronger scale awareness.
	%The framework of SA-APN is described as follows.
	First, the SE-DW-correlation results $\mathrm{\textbf{R}}_\mathrm{d}$ from deeper layers are acquired and expanded with a scale dimension by scale-equivariant convolution, so as the results of depthwise 
	%(DW) 
	correlation $\textbf{R}_\mathrm{s}$ from the shallower layers. Second, scale-channel cross-attention $\mathbf{A}_\mathrm{apn}$ and concatenation $\mathbf{C}_\mathrm{apn}$ are calculated respectively. By adding weights to each item, refined features $\mathbf{F}_\mathrm{apn}$ are acquired by the following formula:
	
	\begin{equation}\label{eq7}
		\mathbf{F}_\mathrm{apn}=\mathbf{R}_\mathrm{d} + \lambda_1\mathbf{A}_\mathrm{apn} + \lambda_2\mathbf{C}_\mathrm{apn}
		\quad,		
	\end{equation}
	where 
	$\mathbf{C}_\mathrm{apn}$ is the feature concatenation of both shallow and deep correlation results, and $\lambda$ refers to a weight.
	Third, as the refined features are obtained, the features will pass to %will be output to the process in 
	PSA%pairwise scale-channel attention
	, as shown in Fig.~\ref{fig:main}. On the other hand, scale-aware anchors are also acquired by AGN. 
	
	\noindent\textit{\textbf{Remark 4}}: In Eq.~\ref{eq7}, the design of SA-APN is concerned with scale information of features from the last two backbone layers. SC cross-attention focuses on scale information particularly, while concatenation directly combines the information between two layers in general as a supplement.

	%%%%%%%%%%%%%%%%%%%%%%%%%%%%%%%%%%%%%%%%%%%%%%%%%%%%%%%%%%%%%%%%%%%%%%%%%%%%%%%%%
	%%%%%%%%%%%%%%%%%%%%%%%%%%%%%%%%%%    Experiment    %%%%%%%%%%%%%%%%%%%%%%%%%%%%%
	%%%%%%%%%%%%%%%%%%%%%%%%%%%%%%%%%%%%%%%%%%%%%%%%%%%%%%%%%%%%%%%%%%%%%%%%%%%%%%%%%
	%\vspace{-4pt}
	\section{Experiment}
	In this section, the proposed benchmark to evaluate object tracking methods for UAM approaching,~\textit{i.e.}, UAMT100, is first introduced. A comparison between object tracking in UAM approaching and general aerial scenes is described. Details about the evaluation experiments on UAMT100 and the authoritative aerial tracking benchmark, \textit{i.e.}, UAV123@10fps~\cite{UAV123}, are also given. Finally, real-world tests of SiamSA are demonstrated. 
	%	Both the UAMT100 benchmark and the code are available at \url{https://github.com/vision4robotics/SiamSA}.

	\begin{figure}[t]	
		\centering
		\includegraphics[width=0.82\linewidth]{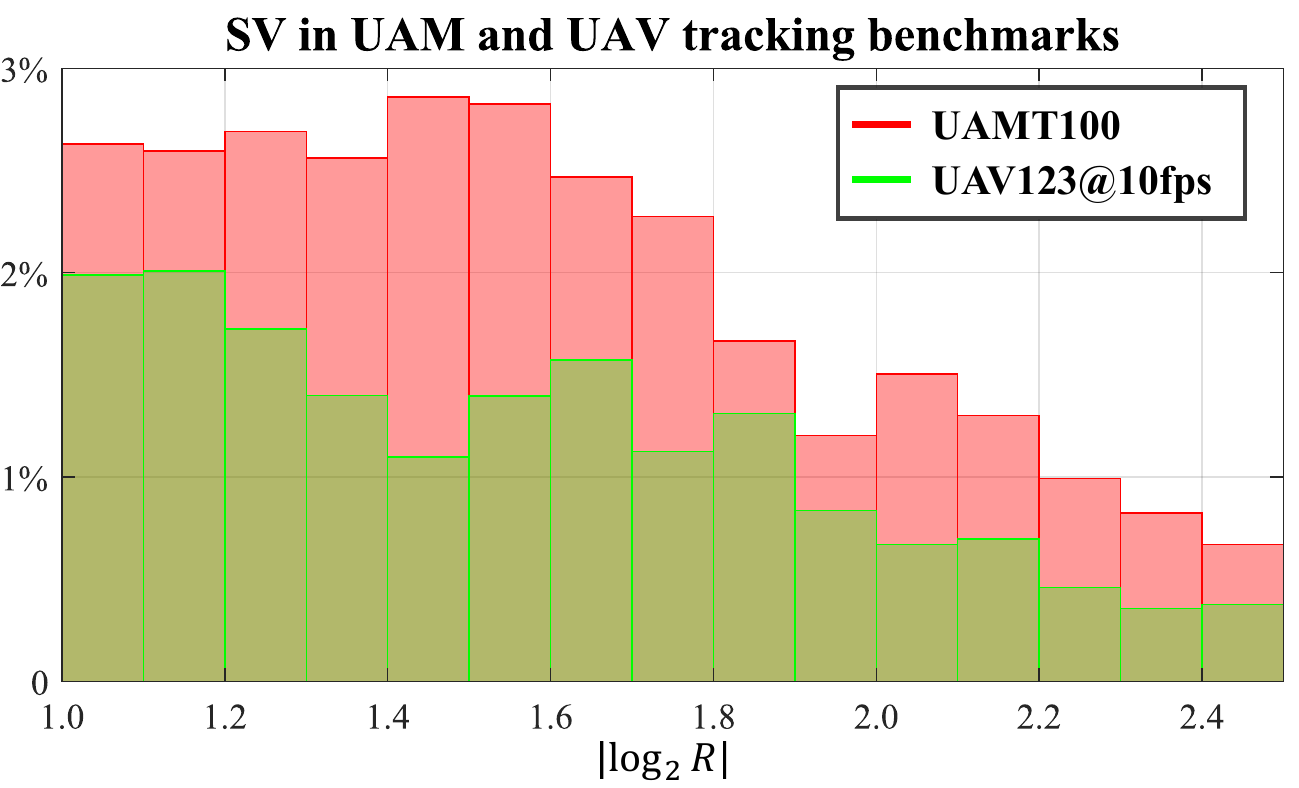}
		\setlength{\abovecaptionskip}{0cm}
		\caption
		{SV comparison plots of UAMT100 and UAV123@10fps. The larger area of the histogram represents more severe SV issues.
		}  
		\label{fig:scatter}
		%			\vspace{-0.05cm}
		\vspace{-0.1cm}
	\end{figure}
	%\vspace{-4pt}
	\subsection{UAMT100 benchmark}
	UAMT100 is recorded to evaluate object tracking methods for UAM approaching.
	The benchmark consists of 100 image sequences, which are captured from a flying UAM platform. 
	It represents various scenarios of UAM's tracking an object for approaching. 
	16 kinds of objects are involved, and 11 attributes are annotated for each sequence.
	The attributes include common challenges of object tracking, \textit{i.e.}, aspect ratio change (ARC), out-of-view (OV), background clutter (BC), fast motion (FM), low illumination (LI), 
	object blur (OB),  partial occlusion (POC), scale variation (SV), similar object (SOB), and viewpoint change (VC). Besides, some challenging scenarios with the UAM attack (UAM-A) by a stick are also considered especially for UAM tracking.  
	The videos are recorded at 10 frames per second (FPS), with the JPG image resolution of 800$\times$600 pixels.
	
	\noindent\textit{\textbf{Remark 5}}: The system to record UAMT100 benchmark is described as follows.
	The OptiTrack\footnote{https://optitrack.com/} Flex 13 camera from Quanser\footnote{https://www.quanser.com/} acquires the pose information of UAM and reports it to the NVIDIA Xavier NX through ROS\footnote{https://www.ros.org/} client nodes for the VRPN library. Besides, the communication between the onboard computer and Pixhawk
	%\footnote{https://docs.px4.io/master/en/flight\_controller/pixhawk.html}
	relies on serial. Moreover, QGroundControl
	%\footnote{http://qgroundcontrol.com/} 
	acts as the ground control station.

	Figure~\ref{fig:scatter} quantitatively demonstrates the difference in SV issue between UAM object tracking for approaching and common UAV tracking. $R$ denotes the degree of SV, which is measured by the ratio of the current object's ground truth bounding box area to the initial one.
	SV is measured when $R$ is outside the range [0.5, 2], \textit{i.e.}, $|\mathrm{log_2}R|>1$.
	The percentage of frames whose $|\mathrm{log_2}R|$ is with a certain section is drawn as the SV histogram, with an interval length of 0.1 over the range of 1 to 2.5. The proportion of sections where $|\mathrm{log_2}R|>2.5$ is less than $0.5\%$ and not of reference significance.
	Therefore, the larger area of the histogram means the higher frequency of object SV, and during UAM tracking for approaching, SV is more common and more severe than in UAV tracking.
	
	\begin{figure}[t]
		%	%\vspace{-6pt}
		\centering	
		{
			\includegraphics[width=0.47\linewidth]{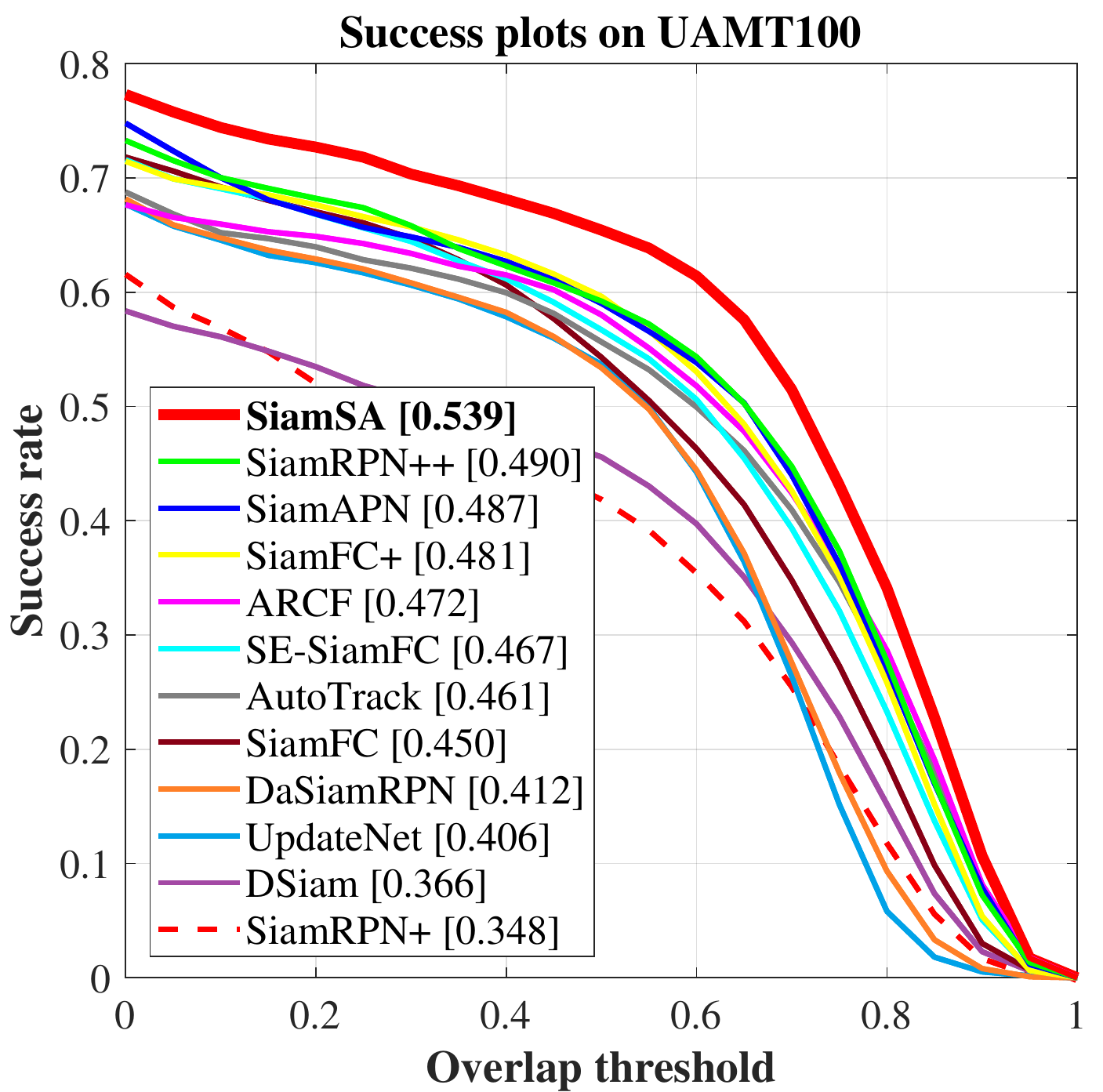}
			\label{fig:UAM}
		}
		{
			\includegraphics[width=0.47\linewidth]{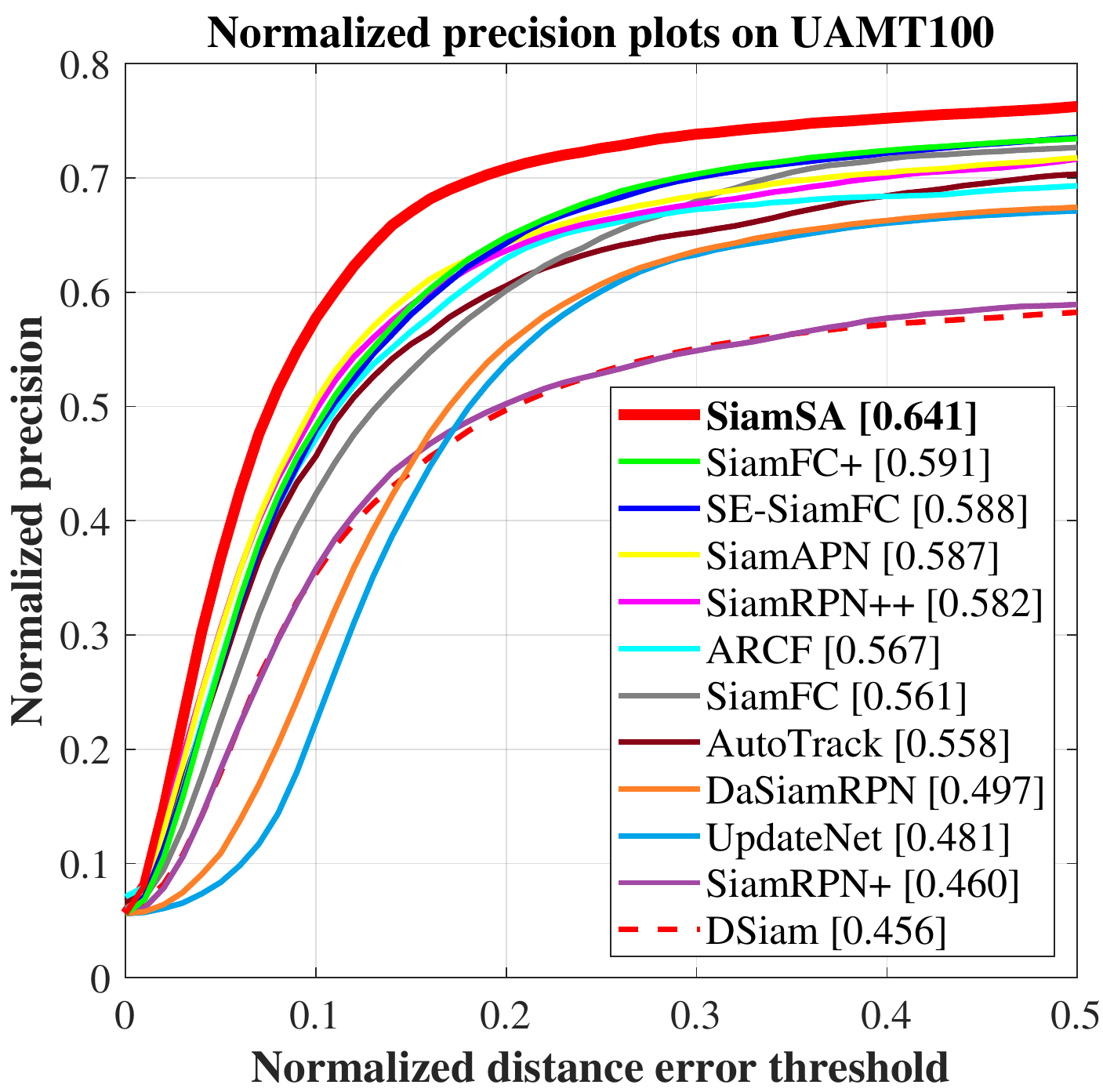}
			\label{}
		}
		%\vspace{-15pt}
		\setlength{\abovecaptionskip}{-4pt}
		\vspace{-0.2cm}
		\caption
		{
			Overall performance evaluation on UAMT100 benchmark. SiamSA surpasses all other trackers in AUC score and NP.
		}
		\label{fig:uam}
		\vspace{-0.1cm}
		%\vspace{-4pt}
	\end{figure}
	%	\begin{figure}[t]	
	%		\centering
	%		\includegraphics[width=0.75\linewidth]{img/radar_UAMT.pdf}
	%		%	\setlength{\abovecaptionskip}{0.15cm}
	%		%\vspace{-6pt}
	%		\caption
	%		{AUC score plots of UAM object tracking attributes on UAMT100. The AUC and NP of SiamSA are shown under the attributes in the format of (AUC, NP), which ranks first in all the attributes.
	%		}  
	%		\label{fig:radar}
	%		%	%\vspace{-0.6cm}
	%		%\vspace{-4pt}
	%	\end{figure}
	\vspace{-4pt}
	\subsection{Implementation details}
	%\vspace{-2pt}
	In SiamSA, The last three layers of the backbone are fine-tuned, and the input size of the template and search image are set to 127×127 and 287×287 pixels respectively. 
	With stochastic gradient descent (SGD), SiamSA is trained on common object tracking training sets, including COCO~\cite{COCO}, ImageNet VID~\cite{VID}, GOT-10k~\cite{Got-10k}, and Youtube-BB~\cite{Youtube}, without extra specific training dataset. 
	All training and evaluating process
	is implemented with an
	Intel i9-9920X CPU, a 32GB RAM, and two NVIDIA TITAN
	RTX GPUs.
	
	%	%\vspace{-4pt}
	\begin{table}[b]
		\centering
		\caption{Comparison with deeper backbones on UAMT100. The best two performances are highlighted with \textbf{\textcolor{red}{red}} and \textbf{\textcolor{blue}{blue}} colors.}
		\vspace{-0.2cm}
		\resizebox{0.86\linewidth}{!}{\begin{tabular}{lccccc}
				\toprule
				\multicolumn{1}{l}{{Trackers}} & {Source} & Backbone & AUC & NP & {FPS}  \\
				\midrule
				Ocean & ECCV 2020 & ResNet50 & 0.418  & 0.525  & {{95}}  \\
				SiamBAN & CVPR 2020 & ResNet50 & 0.494  & 0.582  & 67     \\
				SiamMask & CVPR 2019 & ResNet50 & 0.506  & {{0.605}} & 72     \\
				SiamFC++ & AAAI 2020 & GoogleNet & {{0.521}} & {{0.614}} & \textcolor[rgb]{ 0,  0,  1}{\textbf{106}}  \\
				LightTrack & CVPR 2021 & SuperNet & \textcolor[rgb]{ 0,  0,  1}{\textbf{0.536}} & \textcolor[rgb]{ 0,  0,  1}{\textbf{0.624}} & 56     \\
				\midrule
				\textbf{SiamSA} & \textbf{Ours} & {AlexNet} & \textcolor[rgb]{ 1,  0,  0}{\textbf{0.539}} & \textcolor[rgb]{ 1,  0,  0}{\textbf{0.641}} & \textcolor[rgb]{ 1,  0,  0}{\textbf{122}}  \\
				\bottomrule
		\end{tabular}}%
		\label{tab:deep}%
	\end{table}%
	%\vspace{-0.1cm}
	\subsection{Evaluation metrics}
	%	%\vspace{-2pt}
	The evaluation complies with the classic standard of one-pass evaluation (OPE)~\cite{OPE}, including success rate and normalized precision (NP)~\cite{TrackingNet}. Success rate reflects intersection over union (IoU) score, while NP is concerned with the percentage of frames where the normalized distance between the estimated and ground truth positions is within a threshold. The area under the curve (AUC) of the plot is used to rank the trackers in the experiment.
	%Since the application of object tracking methods for UAM approaching confronts severe scale variation, 
	Notably, NP is less vulnerable to object scale and can better demonstrate tracking robustness compared with the traditional precision metric. Therefore, NP is adopted for UAM approaching.
	
	%	%\vspace{-4pt}
	
	%\vspace{-0.1cm}
	\subsection{Evaluation on UAMT100 benchmark}
	%	%\vspace{-2pt}
	The evaluation of UAMT100 consists of overall performance, attribute-based performance, comparison with deeper backbones, and ablation study.
	
	\noindent\textit{\textbf{Remark 6}}: To be fair, \cite{SiamRPN++}, \cite{SiamAPN}, and \cite{DaSiamRPN} are equipped with AlexNet~\cite{ALEXNET}.
	% pretrained on ImageNet~\cite{ImageNet}. 
	All parameters of trackers are consistent with the paper, and all trackers are evaluated on the same platform.

	%	%\vspace{-0.05cm}
	\subsubsection{Overall performance}
	As shown in Fig.~\ref{fig:uam}, SiamSA outperforms other 11 SOTA trackers~\cite{SiamRPN++}\cite{SiamAPN}\cite{Autotrack}\cite{ARCF}\cite{scaletrack}\cite{DaSiamRPN}\cite{SiamFC,SiamDW,UpdateNet, DSiam}, including Siamese-based and correlation filter-based ones. Compared with the second-best performance, gains on AUC score and NP are \textbf{10.0\%} and \textbf{8.5\%} respectively, which represents the effectiveness of the proposed PSA %pairwise scale-channel attention
	for UAM tracking.

	%	\begin{figure}[t]	
	%		\centering
	%		\includegraphics[width=0.95\linewidth]{img/para.pdf}
	%		\setlength{\abovecaptionskip}{-0.05cm}
	%		\caption
	%		{ Key parameter analysis of $\omega_3$ on UAMT100 benchmark. SiamSA reaches the best performance when $w_3$ is set to 0.90.
	%		}  
	%		\label{fig:sv}
	%			%\vspace{-0.2cm}
	%	\end{figure}
	%	%\vspace{-0.05cm}
	\subsubsection{Attribute-based performance}
	Figure~\ref{fig:attr} demonstrates the attribute-based comparison of 6 top trackers in UAMT100 benchmark. SiamSA ranks first in all the attributes. In challenges that are important to UAM approaching, \textit{e.g.}, %scale variation, aspect ratio change, and out-of-view
	SV, ARC, and OV, SiamSA is particularly outstanding, surpassing the second place by \textbf{10.6\%}, \textbf{8.9\%}, and \textbf{6.2\%}. Besides, on UAM attack,
	%on attributes especially for UAM tracking, \textit{i.e.}, UAM attack, 
	SiamSA also outperforms the second-best performance by \textbf{3.3\%}. The comparison validates that the proposed PSA %pairwise scale-channel attention 
	is competent for scale variation issues, and SiamSA can handle the challenges that are faced in object tracking for UAM approaching.
	
	\begin{figure}[t]
		%		\centering	
		%		{
		%			\includegraphics[width=0.75\linewidth]{img/overall_UAMT100_overlap_OPE.pdf}
		%			\label{fig:UAM}
		%		}
		%		{
		%			\includegraphics[width=0.75\linewidth]{img/overall_UAMT100_np_OPE.pdf}
		%			\label{}
		%		}
		%		% 	%\vspace{-6pt}
		\setlength{\abovecaptionskip}{-4pt}
		
		\centering
		{
			\includegraphics[width=0.47\linewidth]{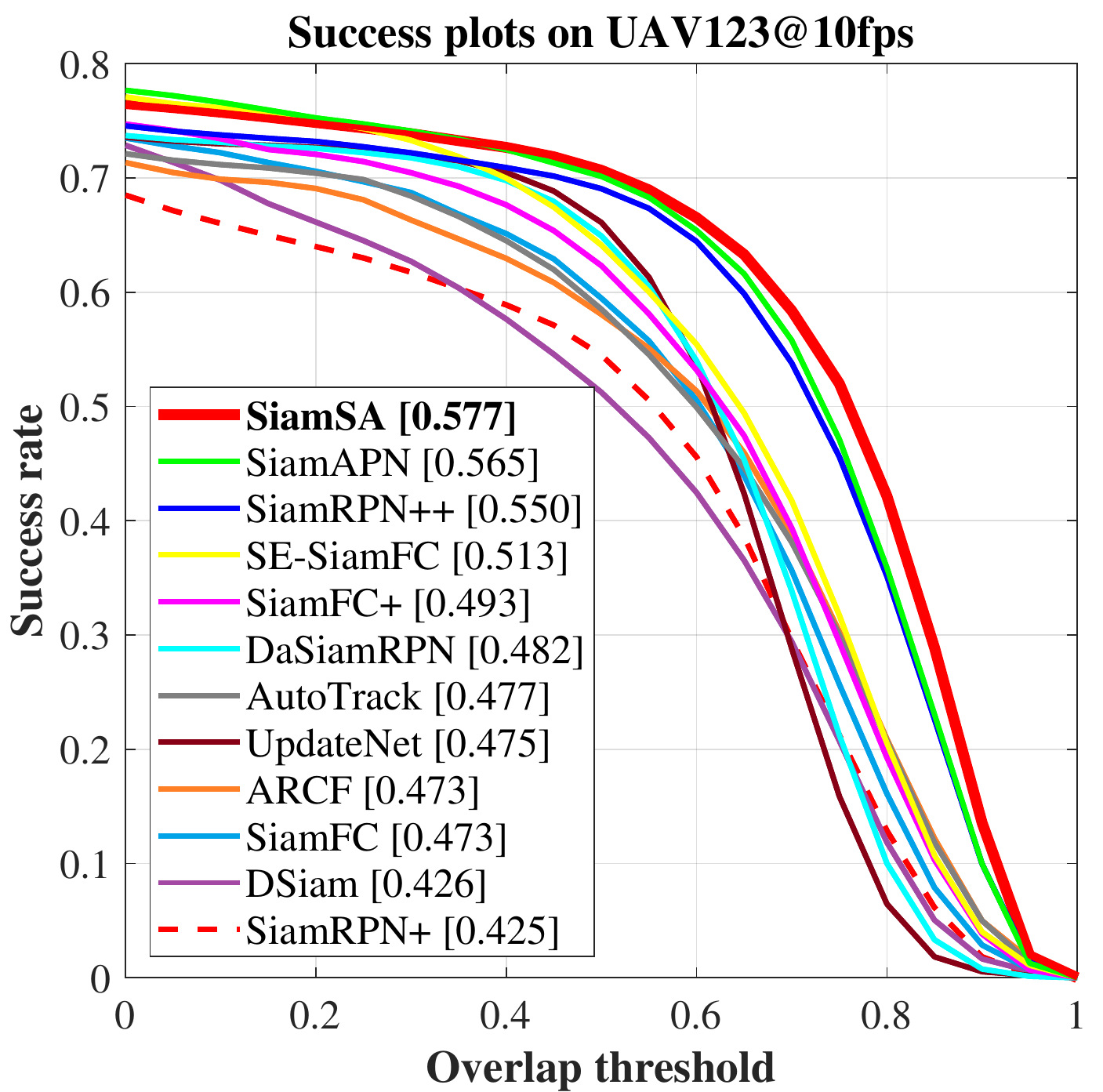}
			\label{}
		}
		{
			\includegraphics[width=0.47\linewidth]{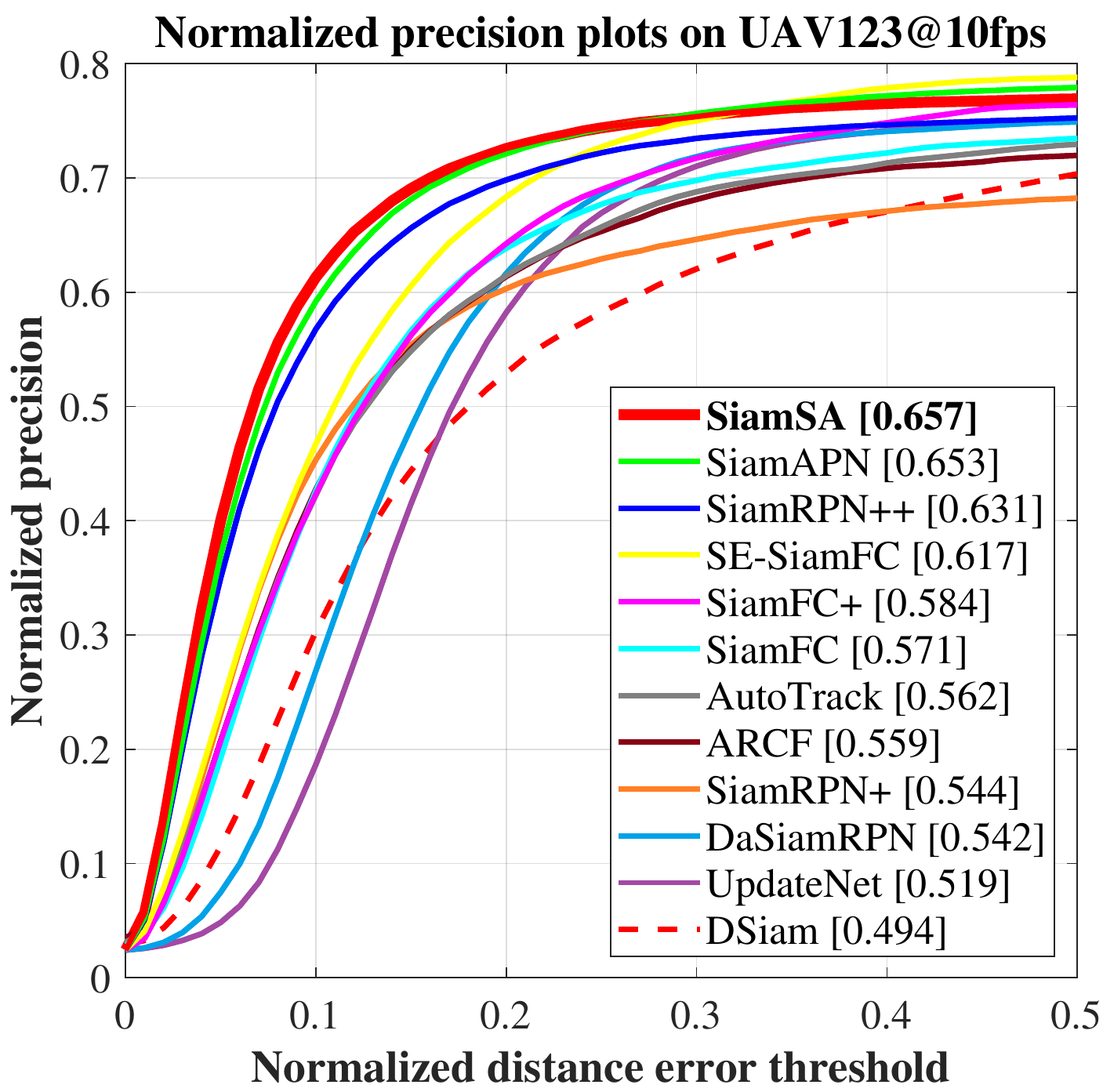}
			\label{}
		}
		%\vspace{-0.4cm}
		\vspace{-0.2cm}
		\caption
		{
			Overall performance evaluation on UAV123@10fps. SiamSA yields competitive performance in terms of AUC score and NP.
		}
		\label{fig:uav}
		\vspace{-0.1cm}
		%\vspace{-0.2cm}
	\end{figure}
	
	\begin{figure}[t]	
		\centering
		\includegraphics[width=0.9\linewidth]{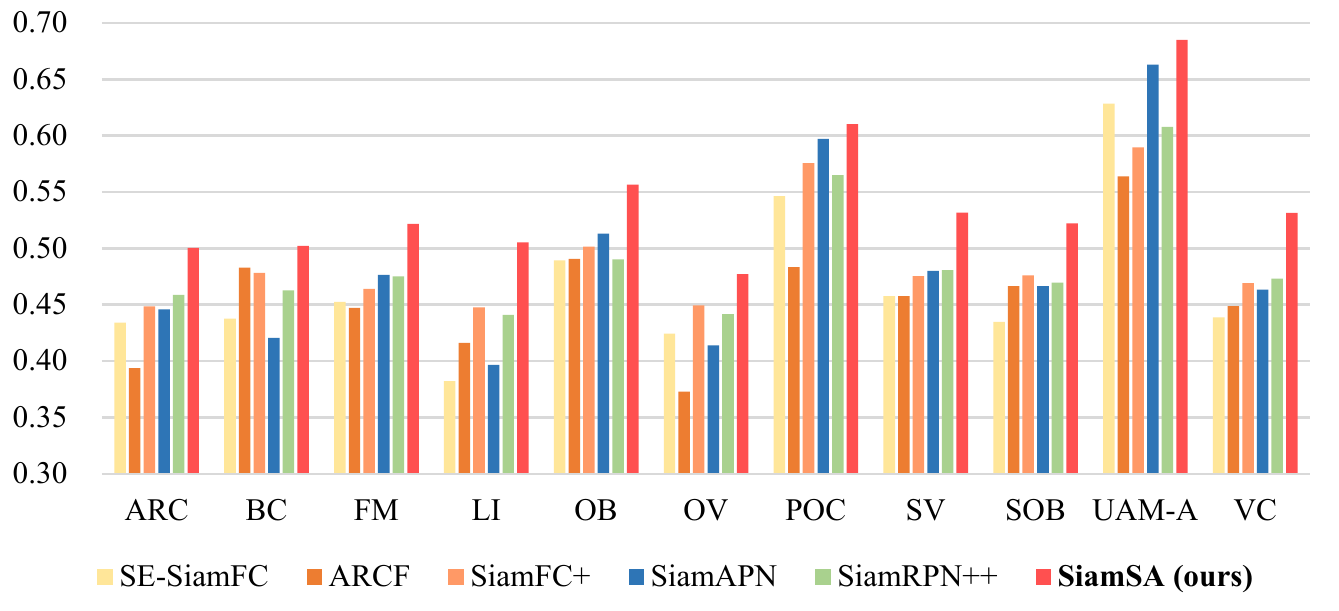}
		\setlength{\abovecaptionskip}{-0.1cm}
		\caption
		{Attribute-based evaluation on UAMT100. SiamSA performs best in all challenges.
		}  
		\label{fig:attr}
		\vspace{-0.1cm}
	\end{figure}

	\begin{table}[b]
		%		%\vspace{-0.6cm}
		\centering
		\caption{Ablation study of SiamSA on UAMT100. $\Delta$ denotes gains.}
		\vspace{-0.2cm}
		\resizebox{\linewidth}{!}{
			\begin{tabular}{lcccc}
				\toprule[0.35mm]
				Trackers & AUC & $\Delta_\mathrm{auc} (\%)$ & NP    & $\Delta_\mathrm{np} (\%)$ \\
				\midrule[0.35mm]
				Baseline & 0.487  & -     & 0.587  & - \\
				Baseline+SA-APN & 0.504  & 3.49\% & 0.610  & 3.91\% \\
				Baseline+PSAN & 0.512  & 5.13\% & 0.625 & 6.47\% \\
				\midrule
				\textbf{Baseline+SA-APN+PSAN (SiamSA)} & \textbf{0.539} & \textbf{10.68\%} & \textbf{0.641} & \textbf{9.20\%} \\
				\bottomrule[0.35mm]
		\end{tabular}}%
		\label{tab:abla}%
		%		%\vspace{-0.6cm}
	\end{table}%
	
	\begin{figure*}[h]	
		\includegraphics[width=0.95\textwidth]{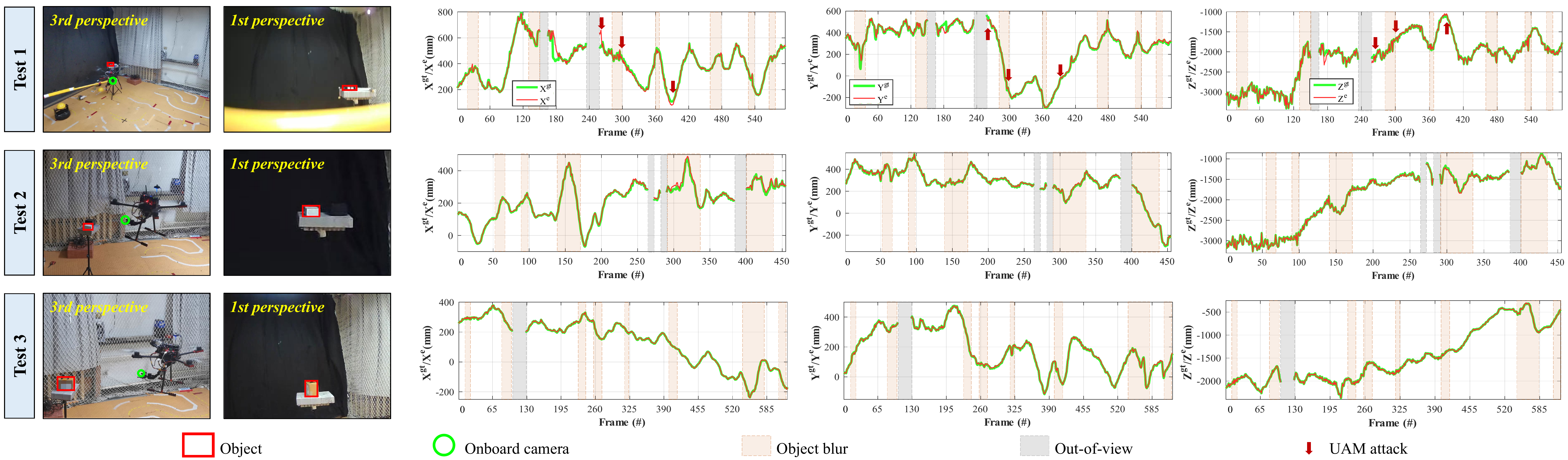}
		\centering
		\setlength{\abovecaptionskip}{0cm}
		\caption
		{
			Real-world vision-based UAM approaching tests with SiamSA for object tracking. The estimated positions X\textsuperscript{e}, Y\textsuperscript{e} and scale S\textsuperscript{e} ({\textcolor{red}{red}} line) is compared with the ground truth X\textsuperscript{gt}, Y\textsuperscript{gt}, and S\textsuperscript{gt} ({\textcolor{green}{green}} line) according to the pinhole camera model. Demo videos of the real-world tests are also available at \url{https://github.com/vision4robotics/SiamSA}.
			%\href{https://github.com/vision4robotics/SiamSA}{here}.
			%			The results show the practicality of SiamSA for vision-based UAM object tracking. The pixel coordinate system, camera coordinate system, and world coordinate system are shown in Fig.~\ref{fig:fig1}.
		}
		\vspace{-0.4cm}
		\label{fig:realworld}
	\end{figure*}
	
	\subsubsection{Comparison with deeper backbones}
	To validate the efficiency of UAM approaching, SiamSA is also compared with 5 SOTA trackers with deeper backbones.
	Notably, deeper backbones generally bring better performance but heavier computational burdens, which can not meet the requirement of real-time tracking. % in UAM approaching. 
	In TABLE~\ref{tab:deep}, despite %the fact that 
	SiamSA is equipped with the lightweight AlexNet, it outperforms other 5 trackers~\cite{SiamFC++}\cite{LightTrack, SiamMask, SiamBAN, Ocean} with SuperNet, ResNet50, or GoogLeNet. Specifically, SiamSA is \textbf{2.18}${\times}$ faster than the second-best tracker, while exceeding the second-fastest tracker by \textbf{3.5\%}. The comparison shows that SiamSA targets well at UAM tracking efficiently.
	%for approaching with high efficiency.

	%	%\vspace{-0.05cm}
	\subsubsection{Ablation study}
	TABLE~\ref{tab:abla} validates the effectiveness of the proposed pairwise scale-channel attention. With only the SA-APN, anchors that adapt to the object scale are generated for object locating, which promotes the success rate by 3.49\%. On the other hand, with PSAN%pairwise scale-channel attention network
	, the feature maps 
	%from the correlation results and APN 
	are equipped with scale awareness, which improves the tracker by 5.13\%. Combing both modules, scale attention works in a pairwise structure, which further excavates the scale information
	to track the object robustly and is more competent for vision-based UAM approaching.

	\begin{table}[b]
		\centering
		\caption{Attribute-based evaluation on UAV123@10fps. The best two results are highlighted with \textbf{\textcolor{red}{red}} and \textbf{\textcolor{blue}{blue}} colors. 
			\vspace{-0.2cm}
			%SiamSA demonstrates excellent generality on attributes that are closely related to UAM tracking.
		}
		\resizebox{\linewidth}{!}{\begin{tabular}{lrrrrrrrr}
				\toprule 
				\multirow{2}[3]{*}{Attribute} & \multicolumn{2}{c}{SV} & \multicolumn{2}{c}{OV} & \multicolumn{2}{c}{ARC} & \multicolumn{2}{c}{FM} \\
				\cmidrule{2-9}          & \multicolumn{1}{c}{AUC} & \multicolumn{1}{c}{NP} & \multicolumn{1}{c}{AUC} & \multicolumn{1}{c}{NP} & \multicolumn{1}{c}{AUC} & \multicolumn{1}{c}{NP} & \multicolumn{1}{c}{AUC} & \multicolumn{1}{c}{NP} \\
				\midrule
				DaSiamRPN & 0.463  & 0.529  & 0.423  & 0.519  & 0.451  & 0.530  & 0.380  & 0.467  \\
				SiamFC+ & 0.460  & 0.554  & 0.425  & 0.554  & 0.429  & 0.534  & 0.377  & 0.490  \\
				SE-SiamFC & 0.483  & 0.590  & 0.453  & {{0.597}} & 0.461  & 0.583  & 0.401  & 0.522  \\
				SiamRPN++ & {{0.522}} & {{0.604}} & {{0.474}} & 0.589  & {{0.498}} & {{0.596}} & {{0.431}} & {{0.532}} \\
				SiamAPN & \textcolor[rgb]{ 0,  0,  1}{\textbf{0.542}} & \textcolor[rgb]{ 0,  0,  1}{\textbf{0.633}} & \textcolor[rgb]{ 0,  0,  1}{\textbf{0.504}} & \textcolor[rgb]{ 0,  0,  1}{\textbf{0.629}} & \textcolor[rgb]{ 0,  0,  1}{\textbf{0.519}} & \textcolor[rgb]{ 1,  0,  0}{\textbf{0.623}} & \textcolor[rgb]{ 0,  0,  1}{\textbf{0.506}} & \textcolor[rgb]{ 1,  0,  0}{\textbf{0.628}} \\
				\midrule
				\textbf{SiamSA (ours)} & \textcolor[rgb]{ 1,  0,  0}{\textbf{0.551}} & \textcolor[rgb]{ 1,  0,  0}{\textbf{0.634}} & \textcolor[rgb]{ 1,  0,  0}{\textbf{0.535}} & \textcolor[rgb]{ 1,  0,  0}{\textbf{0.652}} & \textcolor[rgb]{ 1,  0,  0}{\textbf{0.527}} & \textcolor[rgb]{ 0,  0,  1}{\textbf{{0.620}}} & \textcolor[rgb]{ 1,  0,  0}{\textbf{0.524}} & \textcolor[rgb]{ 0,  0,  1}{\textbf{0.623}} \\
				\bottomrule
		\end{tabular}}%
		\label{tab:uav_att}%
	\end{table}%
	%	%\vspace{-0.05cm}
	%	\subsubsection{Key parameter analysis}
	%	In SiamSA, $\omega_3$ is the weight of the third classification branch, which constrains the distance between the object center and generated anchors. As the SA-APN proposes a certain number of scale-aware adaptive anchors, the location of anchors becomes crucial to address the SV issue. Therefore, $\omega_3$ shows a significant impact on the SA anchor generation. $\omega_3$ is set from 0.700 to 1.070, and SiamSA reaches the best performance when $\omega_3=\ $0.900. When $\omega_3<\ $0.900, anchors far away from the object are not well inhibited, but when $\omega_3>\ $0.900, the constraint becomes severe and adversely impacts the UAM object tracking. Consequently, $\omega_3$ is set to 0.900 in actual deployment.
	\vspace{-0.1cm}
	\subsection{Evaluation on the UAV benchmark}
	To validate the generality of the proposed PSA, SiamSA is also evaluated on UAV123@10fps with 11 SOTA trackers.

	\subsubsection{Overall performance} The overall performance on UAV123@10fps~\cite{UAV123} is shown in Fig.~\ref{fig:uav}. %UAV123@10fps benchmark contains 123 sequences and 38,000 frames with a frame rate of 10 FPS. 
	The benchmark contains SV and other challenges from aerial perspectives, which provides a reference for the generality of the proposed PSA.
	%Due to a reduction of frame rate, severe appearance variation and position movement of objects are similar to the UAM tracking scenarios. 
	%Therefore, UAV123@10fps provides a reference for UAM tracking and is employed for evaluation.
	Compared with the second-best performance, SiamSA improves \textbf{2.1\%} on AUC score attributed to PSA.

	\subsubsection{Attribute-based performance}
	Various challenges in UAV benchmarks have reference significance for analyzing UAM tracking. Especially, SV, ARC, FM, and OV,  which also saliently influence UAM tracking, are discussed. Performance on these attributes is displayed in TABLE~\ref{tab:uav_att}. PSA contributes to the promotion of SiamSA tracker on the SV issue, with an increase of \textbf{1.7\%} on AUC score. Compared with baseline, AUC score increase on ARC, CM, and OV are \textbf{1.5\%}, \textbf{3.6\%}, and \textbf{6.2\%}, which validate the effectiveness and generality of SiamSA against these challenges.

	%%%%%%%%%%%%%%%%%%%%%%%%%%%%%%%%%%%%%%%%%%%%%%%%%%%%%%%%%%%%%%%%%%%%%%%%%%%%%%%%%
	%%%%%%%%%%%%%%%%%%%%%%%%%%%%%%%%%% Real-World Tests %%%%%%%%%%%%%%%%%%%%%%%%%%%%%
	%%%%%%%%%%%%%%%%%%%%%%%%%%W%%%%%%%%%%%%%%%%%%%%%%%%%%%%%%%%%%%%%%%%%%%%%%%%%%%%%%%
	\vspace{-2pt}
	\section{Real-World UAM Approaching Tests}
	\label{sec:Real-WorldTests}
	Real-world tests of SiamSA on three UAM approaching scenes are shown in Fig.~\ref{fig:realworld}. 
	%CLE is used to value the tracking performance. Based on the indoor UAM tracking system mentioned earlier, SiamSA reaches robust tracking performance with a real-time speed of 10.4 fps on NVIDIA Xavier NX platform. 
	In these tests, SiamSA tracks the object robustly and accurately with over 10 FPS on a UAM, which is equipped with an NVIDIA Jetson Xavier NX and an onboard camera.
	TensorRT
	%	\footnote{https://developer.nvidia.com/tensorrt} 
	acceleration is performed locally. 
	In test 1, a yellow stick is employed to attack the UAM three times during approaching a battery. The attacks generally damage the tracking stability by causing sudden deviations in the tracking results. % between frames. 
	Test 2 confronts periods of UAM fast motion, which bring severe object blur and negatively impact the object perception. % during UAM approaching. 
	In test 3, the object out-of-view also poses a formidable challenge for stable tracking.
	Notably, all the tests suffer from severe object scale variation during the UAM approaching the object. But SiamSA manages to handle these issues and represents strong robustness.% owing to the proposed PSA. 
	
	\noindent\textit{\textbf{Remark 7}}: Three real-world vision-based UAM approaching tests of SiamSA further prove the practicality and effectiveness of SiamSA in practical UAM tracking.

	\vspace{-1pt}
	\section{Conclusion}
	%\vspace{-1pt}
	
	In this work, 
	a Siamese network with pairwise scale-channel attention (SiamSA) is proposed especially for object tracking in vision-based UAM approaching. 
	Specifically, to address the severe scale variation issues in UAM approaching, the novel pairwise scale-channel attention (PSA) is proposed. Based on PSA, a pairwise scale-channel attention network and a scale-aware anchor proposal network are designed. The former mainly refines image features with scale information, while the latter proposes anchors that adapt to the object scale. To fairly evaluate object tracking methods for vision-based UAM approaching, UAMT100 benchmark is recorded with a flying UAM platform. Comprehensive experiments on the proposed UAM tracking benchmark and the authoritative UAV tracking benchmark have validated the effectiveness, efficiency, and generality of SiamSA. Furthermore, the real-world tests with a flying UAM and an onboard camera also prove the practicality of SiamSA for various applications. 
	It is convinced that both SiamSA and the UAMT100 benchmark will facilitate the improvement of vision-based UAM-approaching-related applications. 
	
	%%%%%%%%%%%%%%%%%%%%%%%%%%%%%%%%%%%%%%%%%%%%%%%%%%%%%%%%%%%%%%%%%%%%%%%%%%%%%%%%%
	%%%%%%%%%%%%%%%%%%%%%%%%%%%%%% Acknowledgment %%%%%%%%%%%%%%%%%%%%%%%%%%%%%%%%%%%
	%%%%%%%%%%%%%%%%%%%%%%%%%%%%%%%%%%%%%%%%%%%%%%%%%%%%%%%%%%%%%%%%%%%%%%%%%%%%%%%%%
	\vspace{-2pt}
	\section*{Acknowledgment}
	
	This work is supported by the National Natural Science Foundation of China (No. 62173249), the Natural Science Foundation of Shanghai (No. 20ZR1460100), and the Key R\&D Program of Sichuan Province (No. 2020YFSY0004). 
	
	%\newpage
	%%%%%%%%%%%%%%%%%%%%%%%%%%%%%%%%%%%%%%%%%%%%%%%%%%%%%%%%%%%%%%%%%%%%%%%%%%%%%%%%%
	%%%%%%%%%%%%%%%%%%%%%%%%%%%%%%%%%% Reference %%%%%%%%%%%%%%%%%%%%%%%%%%%%%%%%%%%%
	%%%%%%%%%%%%%%%%%%%%%%%%%%%%%%%%%%%%%%%%%%%%%%%%%%%%%%%%%%%%%%%%%%%%%%%%%%%%%%%%%
	
	%%%%%%%%%%%%%%%%%%%%%%%%%%%%%%%%%%% Reference %%%%%%%%%%%%%%%%%%%%%%%%%%%%%%%%%%%%%%%%%%%%%
	\bibliographystyle{IEEEtran}
	\normalem
	\balance
	\bibliography{root.bib}

\end{document}